\documentclass{article} 
\usepackage[preprint]{colm2026_conference}

\usepackage{microtype}
\usepackage{hyperref}
\usepackage{url}
\usepackage{booktabs}
\usepackage{graphicx}
 \usepackage{amsmath}
 \usepackage{algorithm}
\usepackage{algorithmic}
\usepackage{multirow} 
\usepackage[most]{tcolorbox}
\usepackage[table]{xcolor}
\usepackage{tabularx}
\usepackage{newfloat}
\usepackage{listings}
\usepackage{natbib}
\usepackage{caption}
\usepackage{array}
\usepackage{wrapfig}
\usepackage[table]{xcolor} 

\usepackage{lineno}

\definecolor{darkblue}{rgb}{0, 0, 0.5}
\hypersetup{colorlinks=true, citecolor=darkblue, linkcolor=darkblue, urlcolor=darkblue}

\title{MERMAID: \underline{M}emory-\underline{E}nhanced \underline{R}etrieval and Reasoning with \underline{M}ulti-\underline{A}gent \underline{I}terative Knowle\underline{d}ge Grounding for Veracity Assessment}

\author{Yupeng Cao, Chengyang He, Yangyang Yu, Ping Wang, K.P. Subbalakshmi \\
Stevens Institute of Technology\\
\texttt{\{yca33, che14, yyu44, pwang44, ksubbala\}@stevens.edu}
}

\newcommand{\ourmethod}{\textit{MERMAID }}

\definecolor{oursbg}{RGB}{235,247,242} 
\definecolor{bestblue}{RGB}{26,77,167} 
\definecolor{secondorange}{RGB}{217,108,0} 

\newcommand{\best}[1]{\textbf{\textcolor{bestblue}{#1}}}
\newcommand{\second}[1]{\textbf{\textcolor{secondorange}{#1}}}
\begin{document}

\ifcolmsubmission
\linenumbers
\fi

\maketitle

\begin{abstract}

Assessing the veracity of online content has become increasingly critical. Large language models (LLMs) have recently enabled substantial progress in automated veracity assessment, including automated fact-checking and claim verification systems. Typical veracity assessment pipelines break down complex claims into sub-claims, retrieve external evidence, and then apply LLM reasoning to assess veracity. However, existing methods often treat evidence retrieval as a static, isolated step and limit LLM's ability to iteratively refine retrieval during reasoning or to utilize previously acquired knowledge. In this work, we introduce \ourmethod, a memory-enhanced multi-agent framework that operationalizes agentic thinking for veracity assessment by tightly coupling retrieval with iterative reasoning. A Decomposer agent produces structured cues, and an Executor agent alternates between reasoning steps and actions to gather targeted evidence. Crucially, retrieved evidence is consolidated into a long-term memory and selectively recalled for subsequent, related claims, enabling cross-claim evidence reuse and reducing redundant retrieval. Across three fact-checking benchmarks and two claim verification datasets, evaluated with multiple LLM families (GPT, LLaMA, Qwen), \ourmethod delivers state-of-the-art accuracy while improving search efficiency, highlighting the effectiveness of agentic retrieval–reasoning coupled with long-term evidence memory.

\end{abstract}

\section{Introduction}

Veracity assessment aims to check the veracity of claims disseminated in online content, a task that is increasingly crucial in today’s era of information overload~\cite{lazer2018science, guo2022survey, augenstein2024factuality}. However, veracity assessment is a labor-intensive and time-consuming task that requires retrieving relevant evidence and verifying claims based on that information~\cite{nakov2021automated}. Consequently, this has motivated significant research into automated veracity assessment systems.


\begin{wrapfigure}{r}{.45\textwidth}
    \centering
    \includegraphics[width=\linewidth]
    {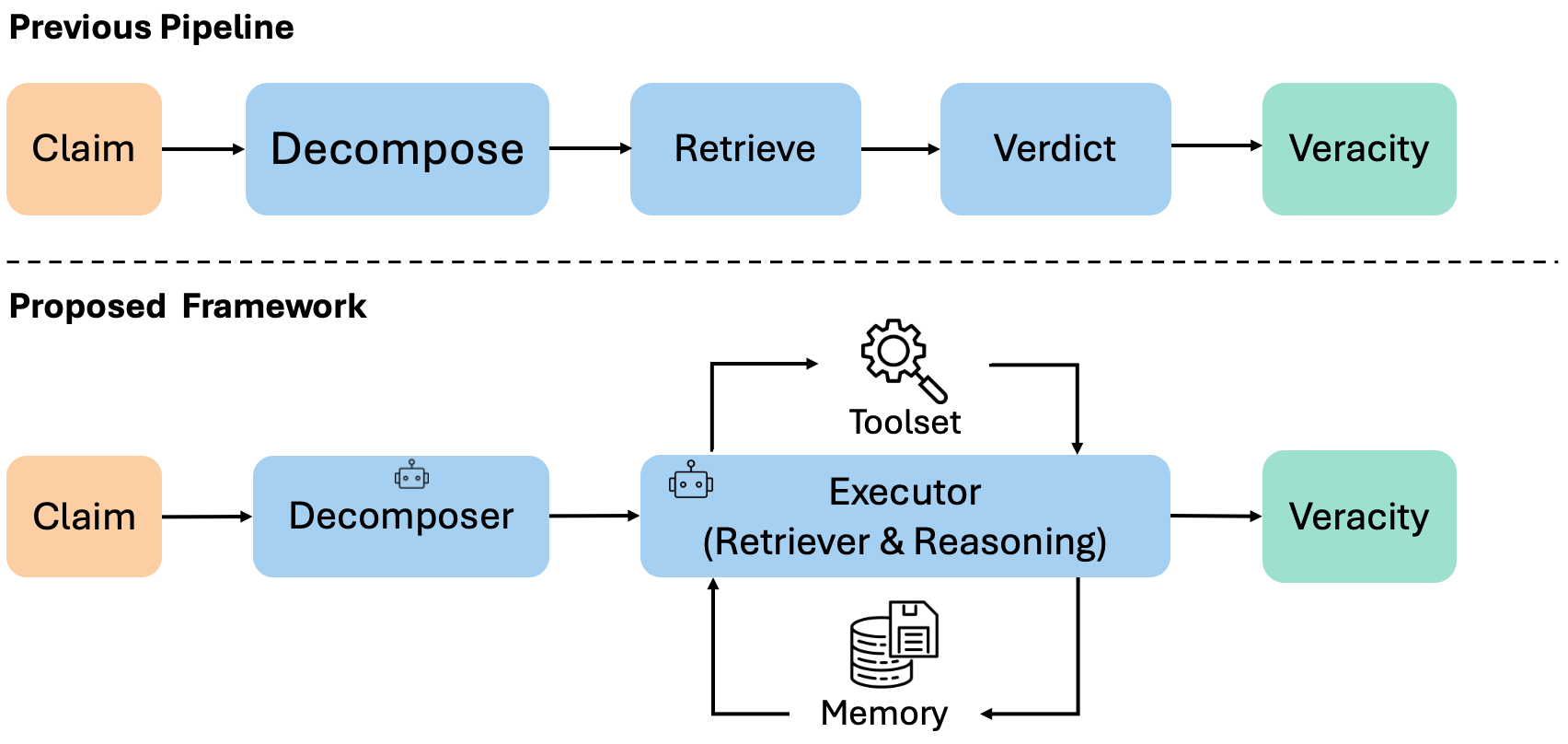}
    \caption{Overall comparison between our framework and existing approaches.}
    \label{fig:previous}
\end{wrapfigure}

Recent advances in language models, particularly their reasoning and agentic capabilities, have created new opportunities for automating veracity assessment. Existing methods typically comprise three components: claim decomposition (or sub-question generation), external evidence retrieval, and reasoning result integration~\cite{chen2022combating, ousidhoum2022varifocal, pan2023fact, wang2023explainable, chern2023factool}. Building on this paradigm, several notable studies use LLMs to develop agent frameworks for veracity assessment. For instance, \citet{zhao2024pacar} leverage the planning capabilities of LLMs to formulate a verification plan before conducting fact-checking. \citet{xie2025fire} and \citet{ma2025local} introduce LLM-based agents capable of conducting multi-turn searches and performing more advanced reasoning processes to verify complex claims.



Despite these advances, existing veracity assessment frameworks still exhibit substantial room for improvement in both reliable accuracy and practical efficiency. A closer look suggests that the bottleneck is often not the capability of LLM-based agents, but how they are instructed to how they search for and consolidate evidence. Many existing pipelines still rely on coarse strategies for tool use and evidence integration, under-utilizing the structured and iterative reasoning that is central to rigorous fact-checking and claim-verification~\cite{shaar2022assisting, warren2025show}. These limitations give rise to two key issues: retrieval is frequently treated as an external, largely stateless component that is only loosely coupled to the agent’s reasoning step, which prevents the agent from iteratively refining queries based on what has already been observed~\cite{yao2022react}.  In contrast, integrating retrieval with the agent’s reasoning modules would allow it to iteratively perform agentic thinking over retrieved evidence and refine queries based on search history, enabling more targeted evidence gathering and more robust veracity assessment. Second, current systems rarely reuse the acquired evidence knowledge in long-term verification tasks. Even when multiple claims share the same entities, events, or underlying factual context, pipelines often perform isolated searches for each claim, overlooking logical dependencies and shared evidence retrieval experience~\cite{panchendrarajan2025multiclaimnet}. This not only wastes retrieval cost, but also makes verification brittle: unlike human fact-checkers who typically recall relevant prior knowledge first and seek additional information only when needed~\cite{weis2023overreliance, finley2023strategic}, LLM pipelines may repeatedly retrieve redundant snippets or inject irrelevant evidence, amplifying noise and inflating context.

To address these challenges, we propose \ourmethod, a memory-enhanced multi-agent framework that jointly performs retrieval and reasoning for veracity assessment (Figure~\ref{fig:previous}). \ourmethod\ tightly integrates retrieval into the agent's reasoning loop, enabling more effective agentic thinking, and introduces a persistent evidence memory module that allows the agent to leverage previously acquired knowledge. The framework orchestrates specialized LLM agents with a structured memory module and a modular, protocol-based toolset that standardizes communication between agents and external knowledge sources. This tool-use design is domain-agnostic: new verification tools can be integrated by wrapping them in a compliant specification, making \ourmethod\ readily applicable to veracity assessment tasks across diverse domains. Specifically, a Decomposer parses each claim into structured representations (e.g., lightweight triplets and topical
context) that condition subsequent retrieval and reasoning. An Executor then performs interactive, step-by-step reasoning within an iterative search loop to gather evidence and determine veracity. Crucially, retrieved evidence is written to long-term memory and can be selectively recalled for related future claims. The memory module adopts an entity-anchored hybrid recall strategy that first performs lexical matching and then applies semantic similarity re-ranking, improving coverage while reducing redundant retrieval steps. By retaining and recalling evidence, \ourmethod\ enables more efficient verification. Finally, the explicit agent trajectories provide transparency and interpretability
compared to pipelines that expose only a final prediction.

We evaluate \ourmethod using several state-of-the-art LLMs, including the GPT, LLaMA, and Qwen families, on five datasets spanning diverse domains and claim types. Experimental results demonstrate that \ourmethod\ achieves state-of-the-art performance on all datasets. Furthermore, the Memory module effectively reduces the number of search operations. In summary, our main contributions are:
\begin{itemize}
    \item We present \ourmethod, a multi-agent framework for veracity
    assessment that tightly couples retrieval with reasoning and
    incorporates a structured memory module for long-term evidence
    knowledge recall, supported by a domain-agnostic toolset that generalizes across verification tasks.
    \item We demonstrate that \ourmethod\ outperforms existing methods on
    five diverse fact-checking and claim verification datasets, with
    consistent improvements even when controlling for the backbone
    model, and that its memory mechanism meaningfully improves search
    efficiency.
    \item We conduct ablation studies on agent configurations and
    evidence retrieval behaviors, and provide an interpretable case
    study of the agentic reasoning process that also uncovers
    annotation issues in existing datasets.
\end{itemize}

\section{The Proposed \ourmethod\ Framework}
\label{sec:method}

\begin{figure*}[t]
    \centering
    \includegraphics[width=\textwidth]{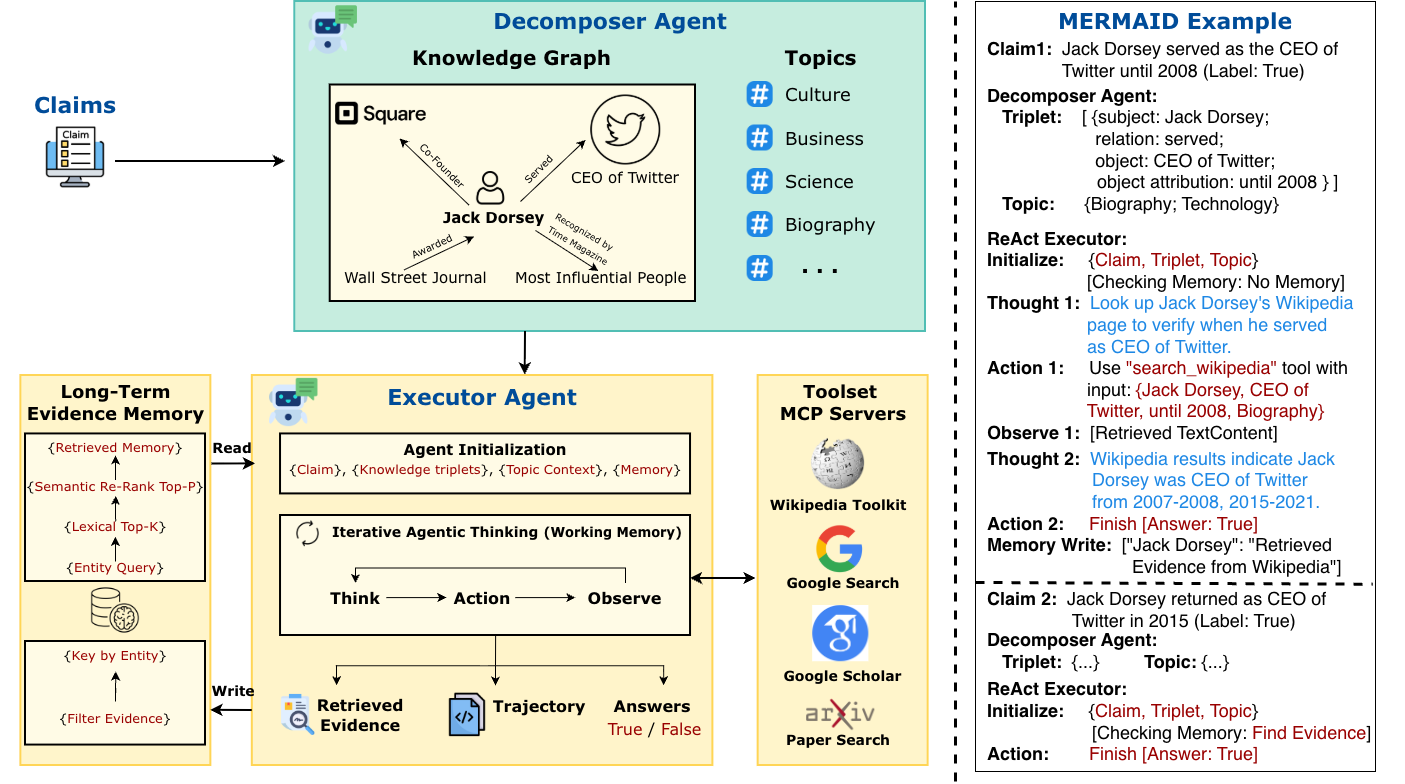}
    \caption{\textbf{Left)} Overview of the MERMAID architecture. The framework comprises a Decomposer agent, an Executor agent, a Toolset, and a Memory module. The Decomposer agent transforms the input claim into a structured knowledge graph and topical context. The Executor agent then engages in an iterative loop, using the Toolset to retrieve evidence and evaluate the claim’s veracity. The retrieved evidence is stored in the Memory module. \textbf{Right)} An example of MERMAID Workflow.}
    \label{fig:workflow}
\end{figure*}

Given an unverified claim, the goal is to determine its veracity by leveraging both newly fetched evidence and relevant prior evidence.  We \emph{orchestrate} the \ourmethod\ framework into four core modules: (i) a \textit{Decomposer} agent, which maps the input claim into a structured knowledge representation to support subsequent agentic reasoning; (ii) an \textit{Executor} agent, which conducts step-by-step agentic thinking for reasoning and evidence gathering; (iii) an extensible \textit{Toolset}, enabling the Executor to interact with external environment to fetch knowledge from diverse resources; and (iv) a \textit{Memory Module}, which stores and reuses previously retrieved evidence and accumulated experience. Figure~\ref{fig:workflow} illustrates the overall workflow, and we formalize and describe these components in detail below. 

\subsection{Decomposer Agent}
For an input claim $c$, the first stage of verification is to analyze the claim, which provides a blueprint for subsequent verification steps. The Claim Decomposer agent $\phi$ transforms the raw input claim $c$ into a structured knowledge representation $D_c = \phi(c) = (G_c, k_c)$, where $G_c$ denotes a set of rational triplets (subject, relation, object, attributions) and $k_c$ represents the topical keywords indicating the semantic domain of the claim. This step serves two purposes. First, the resulting structured representation $D_c$ serves as the foundation for subsequent retrieval and reasoning. By breaking $c$ into atomic factual assertions and identifying key entities, relations, and domain information, $D_c$ highlights what will be checked. We observed that the extracted structured contexts and topical keywords related to $c$ guide the LLMs to a more effective evidence-retrieval strategy. Second, the extracted subjects and objects serve as indices to construct the evidence memory base $M$ (See Section 3.4). We include the complete prompt setup in the Appendix A.1.

To mitigate error propagation from this step to the subsequent verification process, we manually checked the quality of the triplet generation and topic relevance produced at this stage.  We annotated 200 randomly sampled generated triplets, labeling the subject, relation, object, and attribution with binary “Yes/No” judgments. The observed error rates were 1.5\% for subjects, 4\% for relations, and 7.3\% for objects. While some extracted entities were not exact string matches, they effectively preserved semantic fidelity. Additionally, all 200 topic classifications were deemed relevant. Therefore, these error rates fall within an acceptable range and are unlikely to affect the subsequent search and reasoning stages.

\subsection{Executor Agent}
Following the Decomposer agent, the Executor agent verifies the claim $c$ by interleaving logical reasoning with tool-assisted evidence retrieval.  We adopt the Reasoning-Action (ReAct) as the backbone agent behavior pattern, which supports iterative problem solving via alternating reasoning and evidence acquisition, in a manner analogous to how humans search for information while forming judgments. 

\paragraph{Initialization.}
At $t=0$, we construct the initial prompt $P_0$ by combining the original claim $c$, the decomposition output $D_c$, relevant evidence $M_c$ retrieved from the evidence memory $M$, and an empty chat history $H_0=\emptyset$. Specifically, $M_c$ is retrieved from $M$ using the subject and object entities in $D_c$ as queries. The chat history serves as working memory, storing intermediate results generated during multi-step claim verification. The initial prompt $P_0$ is thus: $P_0 = [c;\, D_c;\, M_c;\, H_0].$

\paragraph{Iterative execution.}
For each step $t=1,2,\ldots$, the agent follows a $\{\text{Thought} \rightarrow \text{Action} \rightarrow \text{Observation}\}$ cycle. The agent's policy $\pi$, parameterized by the backbone LLM, generates a thought $th_t$ and an action $a_t$ conditioned on the current prompt: $(th_t, a_t) = \pi(P_{t-1}).$

If $a_t$ corresponds to a retrieval operation, the agent executes $a_t$ using the toolset $\tau$ and obtains a new observation $o_t = \tau(a_t)$. The chat history is then updated by appending the new tuple: $ H_t = H_{t-1} \oplus (th_t, a_t, o_t)$, where $\oplus$ denotes sequence concatenation. The prompt is updated accordingly as $P_t = [c;\, D_c;\, M_c;\, H_t]$. This iterative process continues until the model determines that the task is complete. 

\paragraph{Termination and output.} A termination condition is reached when the policy selects a special \texttt{Answer} action or when a predefined maximum number of steps $T_{\max}$ is exceeded. If the agent does not select \texttt{Answer} by $t=T_{\max}$, we force it to output a veracity judgment based on the accumulated evidence and reasoning history $H_{T_{\max}}$. The Executor produces the predicted label $\hat{y}$ for the claim, along with a human-interpretable rationale (the reasoning-action trajectory and supporting observations). Finally, the Executor writes the newly retrieved evidence back to the evidence memory $M$. The full procedure is shown in Algorithm~\ref{alg:mermaid} (see in Appendix~\ref{app:algo}) and the whole prompt setup is provided in Appendix~\ref{appendix:prompt1}.

\subsection{Augmented Evidence Memory}
In this part, we explain the details of Long-Term Evidence Memory $M$. To enable the agent to apply the knowledge it has previously learned and to reduce redundant retrieval and exploit cross-claim dependencies, \ourmethod\ maintains a persistent evidence memory that is incrementally updated as the system processes a sequence of claims. This module functions as a dynamic knowledge repository that evolves as the system processes sequential claims. Structurally, we implement $M$ as the memory module where we save the atomic entities (subjects and objects) extracted by the Claim Decomposer, and values comprise retrieved evidence along with source metadata (searching query and timestamp). 

Memory access follows a \emph{Recall-then-Update} process. \textbf{(1) Recall.} Before the Executor starts verifying claim $c$, it queries $M$ using the entity set in $D_c$ and retrieves an evidence pack $M_c$ to warm-start the reasoning process. Then, memory recall uses \textbf{\emph{Entity-anchored Hybrid Retrieval Strategy}} over stored chunks, combining sparse lexical matching and dense semantic similarity to improve coverage while controlling evidence noise. First, we use BM25 to select the top-k most relevant pieces of evidence. Then, we perform sentence-level semantic similarity re-ranking to select the top-p most relevant sentences as the evidence memories retrieved from memory. The final memory evidence is further post-processed with de-duplication, diversity-aware selection, and a strict token budget before being injected into the initial prompt $P_0$. \textbf{(2) Update.} After claim $c$ is verified, the system filters the newly collected evidence (e.g., evidence used by the Executor in the final decision) and commits the retained chunks back to $M$ under their associated entity keys, together with provenance metadata. This incremental update mechanism allows \ourmethod\ to progressively accumulate reusable evidence and improve both consistency and efficiency for future, related claims.

\subsection{Toolset for Evidence Composition}
During the Executor’s reasoning-retrieval verification process, the agent acquires evidence through a predefined set of external tools $\tau$. We implement tool interfaces using the Model Context Protocol (MCP), which standardizes communication between an LLM and external tools. Compared to direct LLM tool-calling implementations, MCP provides a model-agnostic invocation interface with standardized schemas and consistent error handling, which significantly reduces tool-calling failure rates and  hallucinations~\cite{patil2024gorilla, hasan2025model}. \ourmethod\ exploits this design to support modular tool composition. Specifically, our tool server exposes (i) five granular Wikipedia tools for multi-level evidence extraction, (ii) Google Search for broad open-domain queries, (iii) Google Scholar Search for retrieving abstracts of scientific publications, and (iv) a paper search tool for verifying scientific claims from PubMed or arXiv website. When the agent executes an action $a_t$, the interface handles the specific tool logic to generate observation $o_t$. This setup ensures that the inference policy $\pi$ remains generic, independent of low-level implementation details. Implementation details of the tool server are provided in the Appendix A.3. This implementation is also readily extensible, and additional tools can be integrated by wrapping them in an MCP-compliant specification.

\section{Experiment Setup}

\subsection{Datasets}
To evaluate the effectiveness of \ourmethod across diverse veracity assessment scenarios, we conduct experiments on five benchmark datasets encompassing distinct claim types, knowledge domains, and difficulty levels spanning two categories:
\textbf{(1) LLM-Generated Response Fact-Checking:} FacTool-QA~\cite{chern2023factool},
BingCheck~\cite{li2024self}, and FactCheck-Bench~\cite{wang2024factcheck}, which
require decomposing long-form LLM outputs into atomic claims and verifying them
against external evidence. Following FIRE~\cite{xie2025fire}, we consolidate labels
into a binary (True/False) format.
\textbf{(2) Natural Claim Verification:} HoVer~\cite{jiang2020hover}, which requires
multi-hop reasoning over Wikipedia. SciFact~\cite{wadden2022scifact} targets scientific claim verification, and we use this dataset to test the framework’s ability to verify scientific claims. Detailed dataset descriptions are provided in Appendix~\ref{app:datasets}

\subsection{Baselines}
Because our evaluation spans heterogeneous datasets, we select baselines to match
each category. For LLM-Generated Response datasets, we compare with
FacTool-QA~\cite{chern2023factool}, FactCheck-GPT~\cite{wang2024factcheck},
SAFE~\cite{wei2024long}, and FIRE~\cite{xie2025fire}. For Natural Claim datasets,
we compare with Self-Ask~\cite{press2023measuring},
ProgramFC~\cite{pan2023fact}, and FOLK~\cite{wang2023explainable}.
While many of these baselines incorporate tool-assisted retrieval, they treat
evidence acquisition as a separate component rather than tightly coupling it with
reasoning, and generally lack explicit long-term memory for cross-claim evidence
sharing. Detailed baseline descriptions are provided in Appendix~\ref{app:baselines}.

\subsection{Implementation Details}
We investigate several SOTA language models, including both proprietary and open-source models. For proprietary models, we evaluate the GPT family (GPT-4o and GPT-5-mini). For open-source models, we assess performance across multiple parameter scales, including LLaMA-3.1 (8B, 70B), Qwen-2.5 (7B, 70B), and the OpenAI-OSS series (20B, 120B). Models in the 7B–20B parameter range are deployed on dual NVIDIA RTX A6000 GPUs, while larger models (70B and 120B) are accessed via serverless computing services. We set the maximum number of steps, $T_{\max}$, to 5 for the Executor agent. For the memory recall strategy, we set the
BM25 candidate pool size to $K = 3$ and the semantic re-ranking
cutoff to $P = 20$.

\subsection{Evaluation Metrics}
We evaluate methods along two dimensions: \emph{overall performance} and \emph{search efficiency}. For overall performance, we compute class-wise F1 scores for the positive and negative labels and report their unweighted macro average (Macro-F1). Concretely, we use \texttt{True}/\texttt{False} for the LLM-Generated datasets, and \texttt{SUPPORTED}/\texttt{REFUTED} for the Natural Claim datasets. To quantify search efficiency, we measure retrieval cost as the total number of queries issued to external tools during verification.

\section{Main Results}

In this section, we analyze the results on two primary metrics: overall performance and search efficiency (memory module). Given computational constraints, we first utilized three LLM-generated response datasets to optimize our framework's configuration. Based on these results, we then evaluated four selected models (GPT-4o, GPT-5-mini, OSS-120B, and Qwen-2.5-70B) on two Natural Claim datasets. We performed all experiments three times and reported the average.

\subsection{Overall Performance Analysis}
As shown in Table~\ref{tab:main}, \ourmethod\ consistently outperforms existing baselines across FacTool-QA, BingCheck, and FactCheck-Bench. We first compare against the previous baselines using the same backbone with GPT-4o, \ourmethod\ improves Macro-F1 over FIRE on FacTool-QA from 0.77 to 0.80, BingCheck from 0.76 to 0.77, and FactCheck-Bench from 0.76 to 0.77, confirming that the gains stem from our framework design rather than a stronger backbone. Then, the results show that \ourmethod\ with GPT-5-mini further improves overall performance, achieving the highest Macro-F1 on FacTool-QA (0.81), BingCheck (0.80), and FactCheck-Bench (0.79). Notably, on BingCheck, \ourmethod\ attains the highest False-F1 of 0.70 while maintaining the True-F1 of 0.89, indicating effectiveness at detecting false claims under web-evidence settings. On FactCheck-Bench, GPT-5-mini delivers the best
False-F1 and Macro-F1, while multiple \ourmethod\ variants tie for the best True-F1. \textbf{These results highlight \ourmethod's robustness across diverse real-world claim verification scenarios, with consistent improvements observed even when controlling for the backbone model.}

\begin{table*}[h]
\centering
\small
\setlength{\tabcolsep}{4pt}
\renewcommand{\arraystretch}{1.01}
\resizebox{0.99\linewidth}{!}{
\begin{tabular}{llccc@{\hspace{6pt}}ccc@{\hspace{6pt}}ccc}
\toprule
\textbf{Framework} & \textbf{LLM} 
& \multicolumn{3}{c}{\textbf{FacTool-QA}} 
& \multicolumn{3}{c}{\textbf{BingCheck}} 
& \multicolumn{3}{c}{\textbf{FactCheck-Bench}} \\
\cmidrule(lr){3-5} \cmidrule(lr){6-8} \cmidrule(lr){9-11}
& & True F1 & False F1 & Macro-F1
  & True F1 & False F1 & Macro-F1  
  & True F1 & False F1 & Macro-F1  \\
\midrule
FacTool & GPT-4o
& 0.84 & 0.58 & 0.71
& 0.68 & 0.56 & 0.62
& 0.82 & 0.64 & 0.73 \\
FactCheck-GPT & GPT-4o
& 0.84 & 0.60 & 0.72
& 0.77 & 0.59 & 0.64
& 0.83 & 0.65 & 0.74 \\
SAFE & GPT-4o
& 0.88 & 0.63 & 0.76
& \second{\underline{0.87}} & 0.65 & 0.76
& 0.84 & 0.65 & 0.74 \\
FIRE & GPT-4o
& 0.89 & 0.66 & 0.77
& \second{\underline{0.87}} & 0.63 & 0.76
& 0.85 & \second{\underline{0.66}} & 0.76 \\
\midrule

\rowcolor{oursbg}
\multirow[t]{8}{*}{\textbf{\ourmethod}}
& LLaMA-3.1-8B-Inst
& 0.63 & 0.41 & 0.52
& 0.60 & 0.46 & 0.53
& 0.62 & 0.45 & 0.54\\

\rowcolor{oursbg}
& LLaMA-3.1-70B-Inst
& 0.69 & 0.44 & 0.57
& 0.65 & 0.56 & 0.60
& 0.71 & 0.52 & 0.62\\

\rowcolor{oursbg}
& Qwen-2.5-7B
& 0.85 & 0.54 & 0.70
& 0.85 & 0.61 & 0.73
& 0.84 & 0.60 & 0.72 \\

\rowcolor{oursbg}
& Qwen-2.5-70B
& 0.86 & 0.59 & 0.73
& \underline{\second{0.87}} & 0.61 & 0.74
& \underline{\best{0.89}} & 0.61 & 0.75 \\

\rowcolor{oursbg}
& OSS-20B
& 0.86 & 0.54 & 0.70
& 0.84 & 0.64 & 0.74
& 0.81 & 0.53 & 0.67\\

\rowcolor{oursbg}
& OSS-120B
& 0.88 & 0.56 & 0.72
& \best{\underline{0.88}} & \second{\underline{0.68}} & \second{\underline{0.78}}
& 0.87 & 0.62 & 0.74\\

\rowcolor{oursbg}
& GPT-4o
& \best{\underline{0.92}} & \second{\underline{0.68}} & \second{\underline{0.80}}
& \best{\underline{0.88}} & 0.65 & 0.77
& \best{\underline{0.89}} &  0.65 & \second{\underline{0.77}}\\

\rowcolor{oursbg}
& GPT-5-mini
& \second{\underline{0.91}} & \best{\underline{0.71}} & \best{\underline{0.81}}
& \best{\underline{0.89}} & \best{\underline{0.70}} & \best{\underline{0.80}}
& \best{\underline{0.89}} & \best{\underline{0.67}} & \best{\underline{0.79}}\\

\bottomrule
\end{tabular}}
\caption{Performance comparison between different baselines across LLM-Generated Response type datasets. \underline{Blue} = best performance; \underline{Orange} = second-best.}
\label{tab:main}
\end{table*}

Table~\ref{tab:hover-scifact} reports results on HoVer and SciFact. On HoVer, \ourmethod\ achieves the strongest overall performance: GPT-5-mini attains the best F1 on 2-Hop (0.72) and 3-Hop (0.59), while GPT-4o leads on 4-Hop (0.63), yielding gains of +0.01, +0.04, and +0.03 over the respective strongest baselines (ProgramFC, FOLK, FOLK). On SciFact, GPT-4o achieves the best F1 of 0.70, outperforming FOLK by 0.02. These results confirm that \ourmethod\ generalizes well to both multi-hop and scientific verification settings.

\begin{table}[t]
  \centering
  \footnotesize
  \setlength{\tabcolsep}{2pt}
  \renewcommand{\arraystretch}{0.8}

  \begin{tabularx}
  {0.60\linewidth}{
    @{}
    >{\raggedright\arraybackslash}X
    *{4}{>{\centering\arraybackslash}c}
    @{}
  }
    \toprule
    \multirow{2}{*}{\textbf{Framework}}
      & \multicolumn{3}{c}{\textbf{HoVer}}
      & \multirow{2}{*}{\textbf{SciFact}} \\
    \cmidrule(lr){2-4}
      & \textbf{2-Hop} & \textbf{3-Hop} & \textbf{4-Hop} & \\
    \midrule

    Self-Ask
      & 0.54 & 0.49 & 0.52 & 0.61 \\
    ProgramFC
      & \second{\underline{0.71}} & 0.51 & 0.53 & - \\
    FOLK
      & 0.66 & 0.55 & 0.60 & 0.68 \\

    \midrule

    \rowcolor{oursbg}
    \ourmethod\ (Qwen-2.5-70B)
      & 0.65 & 0.52 & 0.57 & 0.66 \\
      \rowcolor{oursbg}
    \ourmethod\ (OSS-120B)
      & 0.68 & 0.50 & 0.60 & 0.65 \\
      \rowcolor{oursbg}
    \ourmethod\ (GPT-4o)
      & \second{\underline{0.71}} & \second{\underline{0.56}} & \best{\underline{0.63}} & \best{\underline{0.70}} \\
      \rowcolor{oursbg}
    \ourmethod (GPT-5 Mini)
      & \best{\underline{0.72}} & \best{\underline{0.59}} & \second{\underline{0.62}} & \second{\underline{0.69}} \\

    \bottomrule
    \end{tabularx}

  \caption{Performance comparison on HoVer and SciFact.}
  \label{tab:hover-scifact}
  \vspace{-0.2in}
\end{table}

Notably, strong performance is not exclusive to the most powerful proprietary models. GPT-4o exhibits consistently robust results, and several open-source models are highly competitive: OSS-120B attains the second-best Macro-F1 on BingCheck with SOTA-level results elsewhere, while Qwen-2.5-70B and even Qwen-2.5-7B deliver stable performance across datasets. \textbf{These findings suggest that \ourmethod\ effectively amplifies the reasoning capabilities of the base model}, \textbf{enabling reliable veracity assessment even with computationally efficient, open-source LLMs.}

We also observe substantially weaker performance from LLaMA-based models. Inspection of their agent working logs reveals repetitive tool-calling patterns: although these models invoke tools actively, excessive evidence gathering inflates context length and degrades downstream reasoning, leading to erroneous predictions despite retrieving correct evidence. We provide further analysis of tool-calling behaviors in the Ablation Study and an interpretability-oriented case study in the Explainability Case Study subsection.

\begin{figure}[b]
    \centering
    \includegraphics[width=1.02\linewidth]{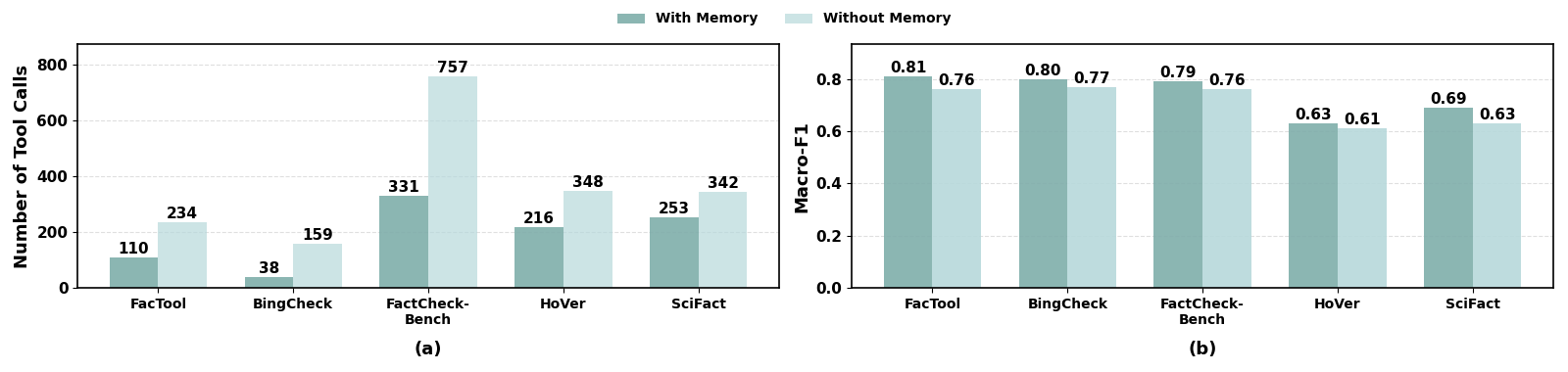}
    \caption{Efficiency and Performance Comparison with vs. without memory (GPT-5-mini).}
    \label{fig:call}
\end{figure}

\subsection{Analysis on Memory Module}
Figure~\ref{fig:call} illustrates the impact of the memory module on both search efficiency and verification performance using GPT-5-mini across all five benchmarks.

\paragraph{Search Efficiency.}
As shown in Figure~\ref{fig:call}(a), enabling memory consistently reduces tool usage across all datasets, lowering the total number of tool calls from 1{,}840 to 948---a reduction of 48.5\%. The most substantial gains appear on BingCheck (159$\to$38, $-$76.1\%) and FactCheck-Bench (757$\to$331, $-$56.3\%), where claims require longer evidence chains and repeated lookups; memory effectively recalls prior results and eliminates duplicative queries.
FacTool also benefits significantly (234$\to$110, $-$53.0\%). On HoVer (348$\to$216, $-$37.9\%) and SciFact (342$\to$253, $-$26.0\%), the relative gains are smaller, likely because the stratified 100-example samples exhibit less topical overlap across claims, leaving fewer opportunities for evidence reuse. Nevertheless, the consistent reductions confirm that \textbf{memory
improves search efficiency across diverse verification settings, with the largest benefits arising in tasks where redundant evidence gathering is most prevalent.}

\paragraph{Verification Performance.}
Beyond efficiency, Figure~\ref{fig:call}(b) shows that memory also
yields consistent performance gains, with Macro-F1 improving across
all five datasets: FacTool (0.76$\to$0.81, +0.05), BingCheck
(0.77$\to$0.80, +0.03), FactCheck-Bench (0.76$\to$0.79, +0.03),
HoVer (0.61$\to$0.63, +0.02), and SciFact (0.63$\to$0.69, +0.06).
The variation in gains complements the efficiency results: SciFact
shows the largest F1 improvement (+0.06) despite the smallest
tool-call reduction ($-$26.0\%), suggesting that memory benefits
SciFact primarily through higher-quality recalled evidence rather
than reduced retrieval volume. These results confirm that
\textbf{the memory module simultaneously enhances efficiency and
effectiveness, making it a key contributor to \ourmethod's overall
performance.}

\section{Ablation Study and Error Analysis}
To evaluate the contribution and effectiveness of individual components within \ourmethod, we conduct several ablation studies. First, we compare the performance of multi-agent versus single-agent configurations. Second, we analyze the agent's evidence retrieval process. Third, we present a case study to show the explainability of the ReAct process.

\subsection{Single-Agent and Multi-Agent}




To study whether explicit decomposition is necessary, we compare a
\emph{multi-agent} setup (Decomposer + Executor) against a
\emph{single-agent} baseline that retains only the Executor, feeding
each claim directly into the Executor loop without intermediate
restructuring. All experiments use GPT-5-mini with $T_{\max}=5$.
Removing the Decomposer yields a consistent Macro-F1 drop across all
benchmarks: FacTool-QA (0.78$\to$0.75), BingCheck (0.79$\to$0.76),
and FactCheck-Bench (0.77$\to$0.74). Without decomposition, the
Executor frequently formulates underspecified retrieval queries,
reducing evidence relevance and wasting limited steps on unproductive searches. The Decomposer mitigates this by structuring claims into clearer verification targets with background context, effectively acting as a lightweight planner that improves retrieval specificity and scaffolds both reasoning and memory construction. This ablation confirms that the Decomposer is not merely auxiliary to memory building; it is critical for \textbf{disambiguating retrieval intents and guiding tool-use under a constrained interaction budget}.

\subsection{Evidence Retrieval Analysis}
Having established the aggregate benefits of the memory module, we now examine the underlying tool-calling behaviors of different LLM backbones and test whether evidence memory transfers across models.
\paragraph{Model Behavior.}
Inspection of ReAct trajectory logs reveals distinct tool-use profiles. LLaMA models frequently fall into repetitive tool-calling loops or experience invocation failures. The OSS series exhibits the opposite tendency, relying heavily on parametric knowledge and invoking external tools only sparingly; within this family, OSS-120B outperforms OSS-20B, reflecting scaling benefits. The GPT and Qwen families strike a more effective balance, using tools selectively and producing higher-quality evidence traces that are easier to consolidate and reuse via memory.

\paragraph{Proxy Evidence Memory.}
In this section, we evaluate whether \emph{established evidence memory} can be reused. Motivated by the last `model behavior' analysis, where LLaMA-3.1-8B-Instruct often struggles with tool-calling and OSS-20B tends to rely on parametric knowledge, we select these two models for targeted testing. We build the evidence memory by running GPT-5-mini on FactCheck-Bench and then reuse this fixed memory with two models: LLaMA-3.1-8B-Instruct (with the MCP toolset disabled, forcing reliance on internal knowledge plus memory) and OSS-20B (with the MCP toolset enabled but memory lookup prioritized). LLaMA-3.1-8B-Instruct improves from 0.54 to 0.65 Macro-F1 (+0.11), and OSS-20B from 0.67 to 0.71 (+0.04). These results indicate that evidence memory transfers effectively across backbone LLMs: it mitigates unreliable tool use in weaker models and provides an external anchor that can correct erroneous parametric beliefs in models that under-utilize retrieval.


\subsection{Memory Recall Strategy}
We compare our hybrid recall strategy against a BM25-only baseline, with both configurations sharing the same memory store and update procedure. The hybrid strategy yields notable gains on GPT-5-mini, the strongest backbone: on BingCheck, Macro-F1 improves from 0.76 to 0.80; on FactCheck-Bench, Macro-F1 increases from 0.77 to 0.79, with both True-F1 and False-F1 also improving. Results on FacTool-QA and with weaker backbones remain largely unchanged. This pattern suggests that hybrid recall primarily benefits datasets with longer evidence chains, where semantically related evidence would be missed by exact matching, and that a sufficiently capable backbone is needed to exploit the richer recalled context.

\subsection{Explainability Case Study and Error Analysis}
\begin{wrapfigure}{r}{.45\textwidth}
    \centering
    \includegraphics[width=\linewidth]
    {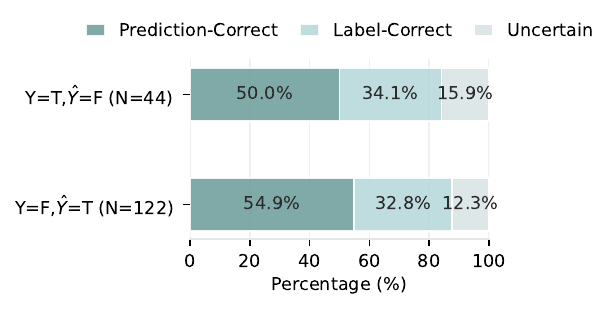}
    \caption{Resolution of label-prediction disagreements ($Y\neq \hat{Y}$), where $Y$ is the gold label, $\hat{Y}$ is the model prediction, and $L$ is the adjudicated label with (True, False, and Uncertain). Bars show \textit{Prediction-Correct}, \textit{Label-Correct}, and \textit{Uncertain}; y-axis lists the disagreement configuration (e.g., $Y=T/F,\,\hat{Y}=T/F$) with group size $N$.}
    \label{fig:disagreement}
    \vspace{-0.15in}
\end{wrapfigure}
The agent trajectory logs produce transparent, step-by-step records
of how the system reaches each veracity decision; a concrete example
in Appendix~\ref{app:case-study} shows how the agent filters
irrelevant recalled memory, triggers fresh search on a cache miss,
and derives the final judgment by comparing the claim's key qualifier against retrieved evidence. 

We also manually reviewed all 166 label prediction disagreements produced by GPT-5-mini across the
three fact-checking datasets, assigning an adjudicated label
$L \in \{\text{T, F, U}\}$. As shown in Figure~\ref{fig:disagreement},
\emph{Prediction-Correct} accounts for the largest share in both
configurations (54.9\% and 50.0\%), suggesting that many mismatches
reflect annotation noise rather than model error. The
\emph{Label-Correct} portion (32.8\%--34.1\%) confirms genuine model
mistakes also occur, and the \emph{Uncertain} segment (12.3\%--15.9\%)
highlights claims that resist binary adjudication even under manual
inspection. Beyond these aggregates, we identify four recurring discrepancy sources:
(1)~direct label errors in the original annotations,
(2)~temporal inconsistency when claims lack explicit time grounding,
(3)~claim quality issues such as missing context or multi-part
statements, and (4)~vagueness and definitional ambiguity in quantities or subjective descriptors. Representative examples are shown in Appendix~\ref{app:errorexample} and Appendix~\ref{app:analysis}.

\section{Conclusions}
In this work, we introduced \ourmethod, a memory-enhanced multi-agent framework that tightly couples retrieval with reasoning for automated veracity assessment. By integrating a ReAct-based Executor with persistent evidence memory, \ourmethod\ enables dynamic evidence gathering and efficient cross-claim reuse. Evaluations across five diverse benchmarks demonstrate state-of-the-art performance with significantly reduced search cost, establishing \ourmethod\ as a scalable and transparent solution for reliable veracity assessment.



\bibliography{colm2026_conference}
\bibliographystyle{colm2026_conference}

\appendix
\onecolumn
\section{Related Work}
\label{app:related}

\subsection{Veracity Assessment Background}
Automated veracity assessment has a long research history, encompassing fact-checking and claim-verification. Earlier studies framed claim veracity assessment as a text classification task using only the claim text as input, where these methods can only capture the surface-level linguistic patterns rather than being based on factual errors~\cite{vlachos2014fact, wang2017liar, perez2018automatic}. Subsequent work introduced evidence-based fact-checking, initially focusing on simple, atomic claims that can be verified with a single piece of `gold' evidence~\cite{thorne2018fever, zhou2019gear, schuster2021get, jiang2021exploring, jin2022towards}. However, complex real-world claims often require multiple pieces of evidence and reasoning. Therefore, recent fact-checking research has established multiple large-scale manually annotated datasets containing complex claims~\cite{aly1feverous, schlichtkrull2023averitec}. Correspondingly, fact-checking methods increasingly emphasize handling complex claims by decomposing raw claims, retrieving external knowledge, and performing multi-step reasoning~\cite{wang2023explainable, chen2024complex}.  While steady progress has been made, current research trends have further expanded to include 1) utilizing LLM to build more agentic veracity assessment workflow~\cite{dmonte2024claim}, 2) addressing factuality challenges posed by large language models, such as verifying LLM-generated responses~\cite{min2023factscore, augenstein2024factuality}, and 3) cross-domain veracity assessment ~\cite{wadden2022scifact, vladika-etal-2024-healthfc, vladika-etal-2025-step}. Our agent framework matches these trends. We feature a structured, modular design with an easily extensible toolset and memory module, allowing seamless integration of new verification tools and the reuse of retrieved evidence. This flexible architecture broadens the system’s applicability to veracity assessment tasks across diverse domains, extending beyond the scope of prior domain-specific approaches.

\subsection{LLM for Veracity Assessment}
The rich knowledge and emergent reasoning capabilities of LLMs present new opportunities for research on veracity assessment. Recently, there have been numerous efforts exploring the use of LLMs for fact-checking and claim verification tasks. Several studies employ different prompting strategies for fact-checking and claim verification tasks, simultaneously validating claims generated by both humans and LLMs~\cite{chen2022combating, pan2023fact, quelle2024perils}. Beyond prompting strategies, researchers have begun developing LLM-based agent frameworks for veracity assessment tasks. \citet{chern2023factool} integrates multiple search tools for detecting factual errors in texts generated by LLMs. \citet{zhao2024pacar} introduces a framework that combines a claim decomposer with self-reflection and a planning module. FIRE~\cite{xie2025fire} proposes an adaptive search procedure to improve search efficiency.  Local~\cite{ma2025local} employs two evaluating agents that iteratively check logical equivalence and causal robustness to detect and correct errors in reasoning. However, these existing methods decouple retrieval from reasoning and lack a persistent memory module for verifying claims over time. Their systems cannot reuse evidence across different claims. Our design addresses this gap by mimicking how human fact-checkers leverage prior knowledge to avoid redundant research on familiar topics.

\clearpage
\section{Algorithm Flow}
\label{app:algo}
\begin{algorithm}[h]
\caption{MERMAID: Joint Retrieval--Reasoning with Evidence Memory for Veracity Assessment}
\label{alg:mermaid}
\small
\begin{algorithmic}[1]
\REQUIRE Evidence memory $M$; Decomposer agent $\phi$; Executor Agent with policy $\pi$; Toolset $\tau$; Max steps $T_{\max}$
\STATE \textbf{Input:} Claim $c$
\STATE \textbf{Output:} Predicted $\hat{y}$; Updated memory $M$; ReAct Trajectory $H$
\STATE \textbf{Decompose:} $D_c \leftarrow \phi(c)$ \hfill // $D_c=(G_c,k_c)$: triplets + topic
\STATE Extract entity keys $\mathcal{E}_c \leftarrow \textsc{Entities}(G_c)$ \hfill // subjects/objects

\STATE \textbf{Executor:}
\STATE Background Context: $D_c$
\STATE Recall Memory: $M_c \leftarrow \textsc{Recall}(M, \mathcal{E}_c)$ \hfill // cached evidence
\STATE Initialize chat history $H_0 \leftarrow \emptyset$
\STATE Initialize prompt/context $P_0 \leftarrow [c;\, D_c;\, M_c;\, H_0]$
\STATE $\hat{y} \leftarrow$ \textsc{None}

\FOR{$t=1$ to $T_{\max}$}
    \STATE $(th_t, a_t) \leftarrow \pi(P_{t-1})$ \hfill // Thought $\rightarrow$ Action
    \IF{$a_t$ is \textsc{Answer}}
        \STATE $\hat{y} \leftarrow \textsc{ParseLabel}(a_t)$
        \STATE \textbf{break}
    \ENDIF
    \IF{$a_t$ is a retrieval/tool action}
        \STATE $o_t \leftarrow \tau(a_t)$ \hfill // new observation
    \ELSE
        \STATE $o_t \leftarrow \emptyset$
    \ENDIF
    \STATE $H_t \leftarrow H_{t-1} \oplus \{(th_t, a_t, o_t)\}$
    \STATE $P_t \leftarrow [c;\, D_c;\, M_c;\, H_t]$
\ENDFOR

\IF{$\hat{y} = $ \textsc{None}}
    \STATE $\hat{y} \leftarrow \textsc{ForceAnswer}(\pi,\,[c;\, D_c;\, M_c;\, H_{T_{\max}}])$
\ENDIF

\STATE \textbf{Update:} $\Delta M \leftarrow \textsc{Evidence}(H_{1:t})$ \hfill 
\STATE $M \leftarrow \textsc{Update}(M, \mathcal{E}_c, \Delta M)$ \hfill // key-value store by entities

\RETURN $\hat{y}, M, H$ \hfill
\end{algorithmic}
\end{algorithm}

\clearpage
\section{Prompts}
\subsection{Decomposer Agent Prompt}
\label{appendix:prompt1}
We show the prompt for the Decomposer Agent below.
\begin{tcolorbox}[colback=gray!5!white,colframe=gray!75!black,title=Decomposer Agent Prompt.,fonttitle=\small,
  fontupper=\small]
  \textbf{System Message:} You are the claim analysis agent in a hierarchical AI Veracity Assessment system. \\

  First, extract the knowledge triplet for the following claim: \#{text}.\\
  The knowledge triplet should consist of these parts: \\
  1. Subject (must be a noun) \\
  2. Relation (must be a verb) \\
  3. Object (must be a noun)\\
  4. Attributes (only use when additional information cannot be captured as a subject, relation or an object) \\

  Second, analyze the topic of the claim.\\
  
  Please provide the answer in JSON format.
  
\end{tcolorbox}

\subsection{Executor Agent Prompt}
We show the prompt for the Executor Agent below.
\begin{tcolorbox}[
  colback=gray!5!white,
  colframe=gray!75!black,
  title=Executor Agent Prompt.,
  fonttitle=\small,
  fontupper=\small
]
\textbf{System Message:} You are a ReAct (Reasoning and Acting) agent for Veracity Assessment.\\[2pt]

Query: \texttt{\{query\}}\\
Background: \texttt{\{background\}}\\
Memory and Previous steps' history: \texttt{\{history\}}\\
Available tools: \texttt{\{tools\}}\\[4pt]

\textbf{CRITICAL:} You MUST respond with ONLY a valid JSON object. No explanations, no markdown, no text before or after the JSON.\\[4pt]

Check the history first.\\
If using a tool:\\
\texttt{\{"thought": "your reasoning", "action": \{"name": "tool\_name", "reason": "why", "input": "query"\}\}}\\[4pt]

If providing final answer:\\
\texttt{\{"thought": "your reasoning", "answer": "True"\}}\\
or\\
\texttt{\{"thought": "your reasoning", "answer": "False"\}}\\[6pt]

Rules:\\
- Output ONLY the JSON object, nothing else\\
- Do NOT wrap JSON in markdown code blocks\\
- Do NOT add any text before or after the JSON\\
- When you have sufficient evidence, provide final answer immediately\\
- Current iteration: this may be your last chance to answer\\
\end{tcolorbox}

\section{Details of Used toolset MCP Servers}

\begin{table}[H]
\centering
\small
\setlength{\tabcolsep}{4pt}
\renewcommand{\arraystretch}{1.08}
\begin{tabularx}{\linewidth}{l l X}
\toprule
\textbf{Server Name} & \textbf{Tool Name} & \textbf{Description} \\
\midrule

\multirow{7}{*}{Wikipedia}
& \texttt{get\_article}              & Get the full content of a Wikipedia article. \\
& \texttt{get\_related\_topics}      & Get topics related to a Wikipedia article. \\
& \texttt{get\_sections}             & Get the sections of a Wikipedia article. \\
& \texttt{get\_summary}              & Get a summary of a Wikipedia article. \\
& \texttt{summarize\_article\_section}& Get a summary of a specific section of a Wikipedia article. \\
& \texttt{search\_wikipedia}         & Search Wikipedia for articles matching a query. \\
& \texttt{extract\_key\_facts}       & Extract key facts from a Wikipedia article. \\
\midrule

Google Search
& \texttt{search\_google}            & Retrieve Google search results for a given query. \\
\midrule

Google Scholar
& \texttt{search\_google\_scholar}   & Retrieve Google Scholar search results for a given query. \\
\midrule

Paper Search
& \texttt{search\_arxiv}             & Search academic papers from arXiv. \\
& \texttt{search\_pubmed}            & Search academic papers from PubMed. \\
\bottomrule
\end{tabularx}
\caption{Toolset overview (MCP servers and tools) used in our framework.}
\label{tab:toolset}
\end{table}

\section{Dataset Details}
\label{app:datasets}
\paragraph{LLM-Generated Response Fact-Checking.}
FacTool-QA~\cite{chern2023factool}, BingCheck~\cite{li2024self}, and
FactCheck-Bench~\cite{wang2024factcheck} focus on verifying long-form responses
generated by LLMs. The system must decompose complex model outputs into checkable
atomic claims and retrieve external evidence for verification. Following
FIRE~\cite{xie2025fire}, we consolidate the original label classifications into
a binary format (True/False).

\paragraph{Natural Claim Verification.}
HoVer~\cite{jiang2020hover} requires multi-hop reasoning over Wikipedia and is
evaluated on 2-hop, 3-hop, and 4-hop subsets. SciFact~\cite{wadden2022scifact}
tests verification of scientific claims against research abstracts. Consistent
with FOLK~\cite{wang2023explainable}, we use the HoVer validation set and apply stratified sampling to select balanced label
distributions. For SciFact, we select only claims with complete evidence
(Support/Refute) to ensure ground-truth labels.

\section{Baseline Descriptions}
\label{app:baselines}

\paragraph{LLM-Generated Response.}
\begin{itemize}
    \item \textbf{FacTool-QA}~\cite{chern2023factool} proposes a tool-augmented
    framework that integrates Google Search to assess the factuality of LLM outputs.
    \item \textbf{FactCheck-GPT}~\cite{wang2024factcheck} targets fine-grained
    factuality evaluation using a benchmark with annotations at the claim, sentence,
    and document levels.
    \item \textbf{SAFE}~\cite{wei2024long} verifies long-form generations by
    decomposing them into atomic facts and checking each via Google Search.
    \item \textbf{FIRE}~\cite{xie2025fire} is an agent-based framework that
    iteratively performs evidence retrieval and fact-checking.
\end{itemize}

\paragraph{Natural Claim.}
\begin{itemize}
    \item \textbf{Self-Ask}~\cite{press2023measuring} decomposes a complex query
    into sub-questions, answers them sequentially, and produces the final answer.
    \item \textbf{ProgramFC}~\cite{pan2023fact} generates an explicit reasoning
    program for each claim and executes it step by step.
    \item \textbf{FOLK}~\cite{wang2023explainable} decomposes a claim into
    sub-claims and performs reasoning over knowledge-grounded QA pairs.
\end{itemize}

\section{Societal Impact Discussion}
\label{appendix:discussion}
Our framework is applicable to a wide range of real-world claim veracity assessment tasks. The MCP-based toolset is designed to flexibly integrate diverse search tools, enabling robust verification across multiple domains. By grounding predictions in evidence memory and exposing interpretable reasoning trajectories, MERMAID reduces verification effort while strengthening trust in automated systems. We also recognize that automated veracity assessment systems may inherit biases from underlying data sources or tools, and should therefore not be treated as authoritative without human oversight. Accordingly, we position MERMAID as a decision-support system that augments, rather than replaces, expert judgment within a responsible information ecosystem.

\section{Examples of Error Analysis}
\label{app:errorexample}
\begin{table}[h]
\centering
\small
\renewcommand{\arraystretch}{1.15}
\resizebox{\columnwidth}{!}{
\begin{tabular}{p{3cm} p{4.5cm} p{5cm}}
\hline
\textbf{Source} & \textbf{Example Claim} & \textbf{Reasoning} \\
\hline
Wrong label 
& \emph{In 1980, Justice William O. Douglas was still alive.}
& The dataset label marks the claim as false, but Douglas died in January 1980; therefore he was alive during part of 1980, making the claim true under a year-level interpretation. \\
\hline
Temporal inconsistency
& \emph{Canadians work an average of 1,702 hours per year.}
& Average annual hours are time-sensitive and can shift across years; reported values also vary by data source and measurement definition. \\
\hline
Claim-quality issue
& \emph{Water exhibits a phenomenon known as 'structural memory.'}
& The term is not operationalized consistently and remains scientifically contested; without specifying the intended mechanism and evidence standard, the claim is difficult to adjudicate with a binary label. \\
\hline
Vagueness / definition ambiguity
& \emph{There are significant Protestant populations in Mexico.}
& ``Significant'' is inherently vague and threshold-dependent; even with an estimated $\sim$12\% share, whether this qualifies as significant depends on the chosen criterion and comparison baseline. \\
\hline
\end{tabular}
}
\caption{Representative discrepancy sources for dataset--model disagreements.}
\label{tab:error-examples}
\end{table}

\section{Example of Explainability Case Study}
\label{app:case-study}
\begin{table*}[h]
\centering
\footnotesize
\setlength{\tabcolsep}{4pt}
\renewcommand{\arraystretch}{1.15}

\resizebox{\textwidth}{!}{%
\begin{tabular}{p{3.2cm} p{13.8cm}}
\toprule
\textbf{Field / Step} & \textbf{Content} \\
\midrule

\multicolumn{2}{l}{\textbf{Stage 1: Claim Understanding (Decomposer Output)}} \\
\midrule
Input Claim &
The cost of making \textit{Pirates of the Caribbean: On Stranger Tides} (2011) was more than \$456M (inflation-adjusted). \\
Gold Label &
False \\
Decomposition (triplet) &
(The cost of making \textit{On Stranger Tides...}, was more than, \$456M inflation-adjusted) \\
Topic &
Entertainment; Economics \\

\midrule
\multicolumn{2}{l}{\textbf{Stage 2: Verification (Executor ReAct Trace)}} \\
\midrule
Init: Memory recall &
Recall cached evidence by entities (film title, budget): (i) unrelated cached item retrieved $\rightarrow$ ignored;
(ii) prior budget query returned no hits $\rightarrow$ proceed to fresh search. \\
Thought 1: & Search for the film's production budget and reliable sources stating its budget and inflation-adjusted figures. \\
Action 1: Search & 
Using \textbf{`search google'} tool with \textit{Query}: ``On Stranger Tides budget ... inflation adjusted \$456M'' \\
& \textit{Observation 1}: mixed cost signals \textbf{(Wikipedia net budget $\sim\$379$M; Forbes cost $\sim\$410.6$M)}. \\
Thought 2: & Search for reliable sources reporting any inflation-adjusted estimates to determine whether it exceeds \$456M. \\
Action 2: Search &
Using \textbf{`search google'} tool with \textit{Query}: ``inflation adjusted ... \$456M source'' \\
& \textit{Observation 2}: multiple sources explicitly state an inflation-adjusted cost of \textbf{\$456M}. \\
Thought 3: & None of the reliable sources indicate it was greater than \$456M after adjustment, so the statement that it was more than \$456M is not supported. \\
Action 3: Answer &
Evidence supports ``\textbf{equals} \$456M'' rather than ``\textbf{more than} \$456M'' $\Rightarrow$ verdict \textbf{False}. \\
\bottomrule
\end{tabular}%
}

\caption{Explainability case study: Stage 1 decomposes the claim into structured context; Stage 2 shows the ReAct-style verification trace leading to the final verdict.}
\label{tab:explain_case}
\vspace{-0.10in}
\end{table*}

In the analysis above, we found that ReAct trajectories yield a transparent, step-by-step record of how the system reaches a veracity decision, making the verification process easy to audit and diagnose. We therefore include a case study to illustrate this interpretability in practice. The table~\ref{tab:explain_case} highlights our two-stage workflow: \textbf{Stage 1} structures the raw claim into a focused verification target (triplet) and a topical scope, which clarifies what evidence is needed; \textbf{Stage 2} then exposes the Executor’s \emph{Thought$\rightarrow$Action$\rightarrow$Observation} loop, including explicit memory checks and the rationale for each retrieval step. In this example, the trace shows (i) first, the agent filters irrelevant recalled memory, (ii) second, a memory miss triggers fresh search, (iii) third, queries are refined to locate inflation-adjusted figures and contents, and (iv) how the final judgment follows from comparing the claim’s key qualifier (``more than'') against evidence supporting ``equals \$456M.'' Overall, these logs provide a faithful, human-readable explanation of both evidence selection and decision-making, strengthening the framework’s interpretability.

\section{Detailed Steps of Error Analysis}
\label{app:analysis}
We conduct an error analysis using the ReAct trajectory logs to better understand why the model makes the wrong decisions. Fig.~\ref{fig:disagreement} summarizes how adjudication resolves cases where the dataset label $Y$ and model prediction $\hat{Y}$ disagree ($Y\neq \hat{Y}$). We focused on three fact-checking datasets. Concretely, we manually reviewed all 166 disagreement cases produced by GPT-4o and assigned an adjudicated label $L \in \{\mathrm{T}, \mathrm{F}, \mathrm{U}\}$ (True/False/Uncertain) based on the claim and the retrieved evidence in the trajectory. We then categorize each case into three outcomes: (i) \textit{Prediction-Correct} if $\hat{Y}=L$, (ii) \textit{Label-Correct} if $Y=L$, and (iii) \textit{Uncertain} if $L=\mathrm{U}$. Notably, the presence of an \textit{Uncertain} segment in both groups shows that some conflicts are not cleanly decidable under binary verification, even with manual inspection.

Across both disagreement configurations, \textit{Prediction-Correct} accounts for the largest share: 54.9\% when $(Y=\mathrm{F}, \hat{Y}=\mathrm{T})$ (N=122) and 50.0\% when $(Y=\mathrm{T}, \hat{Y}=\mathrm{F})$ (N=44). This pattern suggests that a substantial fraction of mismatches are consistent with annotation noise or underspecified labeling guidelines, rather than model error alone. At the same time, the \textit{Label-Correct} portion remains non-trivial in both groups (32.8\% and 34.1\%), indicating that the model also makes genuine mistakes under disagreement settings (e.g., missing key qualifiers or over-trusting weak evidence). Finally, the persistent \textit{Uncertain} segment (12.3\%--15.9\%) highlights that some claims are inherently difficult to adjudicate as strictly True/False without additional context (e.g., a precise timeframe, scope, or operational definition), even after manual inspection.

Beyond these aggregate outcomes, we observed cases in which the model’s reasoning and retrieved evidence were sound, yet the final prediction still disagreed with the gold label. Manual inspection revealed that many of these discrepancies stem from limitations in the datasets rather than model errors. We identify four recurring sources: (1) \textbf{Direct Label Errors:}  Manual adjudication shows that a subset of disagreements arises from incorrect annotations in the original datasets.
(2) \textbf{Temporal inconsistency:} time-sensitive facts may change or be revised (e.g., updated statistics), and when a claim is not explicitly time-grounded, $Y$ and $\hat{Y}$ may implicitly refer to different reference dates; adjudication may thus favor either side, or be marked \textit{Uncertain} if the timeframe cannot be established.
(3) \textbf{Claim quality issues:} include missing context (time, location, population), controversial claims without stable ground truth, and partially correct or multi-part statements; these frequently drive ambiguity because a single binary label cannot faithfully capture the nuanced status of the claim.
(4) \textbf{Vagueness and definitional ambiguity:} approximate quantities (e.g., 46M vs.\ 46.3M), subjective descriptors (e.g., “significant”), and terms with multiple reasonable definitions create boundary cases unless thresholds and definition rules are explicitly specified. We show the detailed examples in Appendix A.4.
\vspace{-0.1in}

\end{document}